\documentclass{article} 
\usepackage{iclr2019_conference,times}

\usepackage{amsmath,amsfonts,bm}

\def\eqref#1{equation~\ref{#1}}

\def\1{\bm{1}}

\DeclareMathAlphabet{\mathsfit}{\encodingdefault}{\sfdefault}{m}{sl}
\SetMathAlphabet{\mathsfit}{bold}{\encodingdefault}{\sfdefault}{bx}{n}

\usepackage{url}

\usepackage{microtype}
\usepackage{graphicx}
\usepackage{subfigure}
\usepackage{booktabs} 
\usepackage{float}

\usepackage{hyperref}

\usepackage[linesnumbered,algoruled,boxed,lined]{algorithm2e}

\usepackage{lscape}
\graphicspath{ {./} }

\title{Hierarchical Policy Learning is Sensitive to Goal Space Design}

\author{Zach Dwiel \\
Intel AI Lab\\
Bloomington, IN, U.S.A.\\
\texttt{zach.dwiel@intel.com} \\
\And
Madhavun Candadai \\
Program in Cognitive Science\\
School of Informatics, Computing and Engineering\\
Bloomington, IN, U.S.A.\\
\texttt{madcanda@indiana.edu} \\
\And
Mariano Phielipp\\
Intel AI Lab\\
Phoenix, AZ, U.S.A.\\
\texttt{mariano.j.phielipp@intel.com} \\
\And
Arjun K. Bansal\\
Intel AI Lab\\
San Diego, CA, U.S.A.\\
\texttt{arjun.bansal@intel.com} \\
}

\iclrfinalcopy 
\begin{document}
\maketitle


\begin{abstract}
Hierarchy in reinforcement learning agents allows for control at multiple time scales yielding improved sample efficiency, the ability to deal with long time horizons and transferability of sub-policies to tasks outside the training distribution. It is often implemented as a master policy providing goals to a sub-policy. Ideally, we would like the goal-spaces to be learned, however, properties of optimal goal spaces still remain unknown and consequently there is no method yet to learn optimal goal spaces. Motivated by this, we systematically analyze how various modifications to the ground-truth goal-space affect learning in hierarchical models with the aim of identifying important properties of optimal goal spaces. Our results show that, while rotation of ground-truth goal spaces and noise had no effect, having additional unnecessary factors significantly impaired learning in hierarchical models.
\end{abstract}

\section{Introduction}
Deep reinforcement learning has recently been successful in many tasks~\citep{Mnih2015_nature}\citep{Silver2017_nature}~\citep{Schulman2015}, but increasing sample efficiency in environments with sparse rewards, performing well in tasks with long time horizons and learning skills transferable to new tasks are still active areas of research~\citep{Nair2018}~\citep{Frans2018}~\citep{Aytar2018}. Hierarchy and unsupervised representation learning are two ingredients that are widely believed to produce significant advances in these domains.


One approach to hierarchical reinforcement learning is Hierarchical Actor-Critic (HAC)~\citep{Levy2017}.
In HAC, higher level policies specify goals for policy levels directly below them, allowing hindsight experience replay (HER)~\citep{Andrychowicz2017} to be integrated with hierarchy.
While this approach has been shown to improve sample efficiency in tasks with sparse rewards, it leaves the designer with two tasks: goal space design and goal success definition.
In this paper we explore goal space design and its effects on learning.
One drawback of the HAC algorithm is that manual design of goal spaces may not be possible in all cases.
Moreover, goal space design can be somewhat arbitrary.
For example, in the pendulum swing up task, goal success is defined as being contained by a bounding box with a manually chosen size surrounding the desired goal and having velocity less than a specific threshold.
The environment itself might come with natural definition of success, but that definition of success is not always the optimal to use for defining success of sub-goals.
Therefore, it is desirable that the goal spaces be also learned and this is where unsupervised representation learning comes in.

\begin{figure}[t]
\centering
\includegraphics[width=0.9\textwidth]{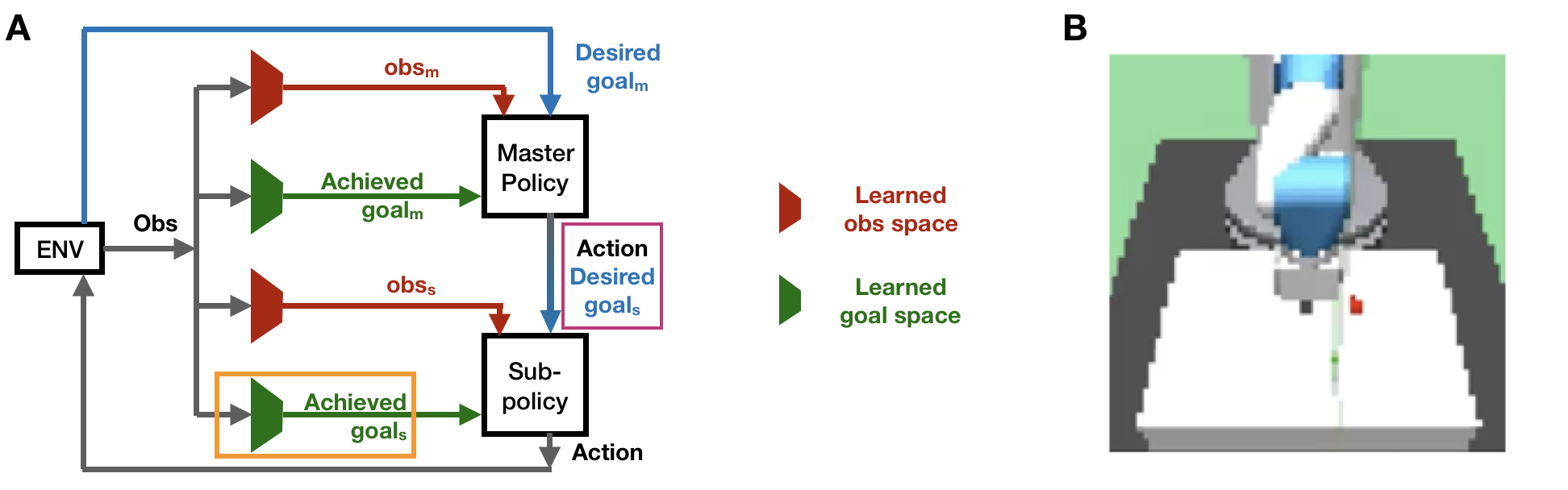}
\caption{Agent architecture and environments. (a) Hierarchical Actor-Critic with two levels, and learned goal and observation spaces. The overall goal of the agent is specified as a goal to the master policy. The subpolicy acts in the environment. The master policy acts by producing goals for the sub-policy. The yellow box denotes the place where modifications are made to the spaces in our analysis, and the entity in the pink box is matched in dimensionality accordingly. (b) We use OpenAI gym’s simulated robotics environment, in particular the Fetch platform~\citep{openai_gym}.}
\label{fig:setup}
\end{figure}

A generic architecture which combines hierarchy, and unsupervised representation learning is shown in figure~\ref{fig:setup}A.
In this architecture, the observation from the environment is mapped onto the internal observation spaces and achieved goal spaces of both the master and sub policies using an auto-encoder.
This learned representation serves as the space in which policy learning happens.
The master policy provides goals to the lower level; meaning, the action space of the higher level policy is the same as the goal space of the lower level policy.
Both observation and goal spaces are not hand-designed but are instead learned in an unsupervised fashion in such an architecture.
However, there are drawbacks that arise - (1) policy learning is subject to the shortcomings of the unsupervised learning methodology that is adopted and (2) the designer still has to identify hyper parameters of the unsupervised representation learning system.
Recently, ~\citep{nair2018visual} presented a model in which they demonstrate superior performance in robotic reach tasks using goal-spaces learned using a variational auto-encoder (VAE), albeit without hierarchy.

While HAC and unsupervised representation learning have been independently shown to be beneficial,
we combine the two by providing intentionally hand-designed goal-spaces that mimic learned goal-spaces.
Despite advances in the literature in learning latent representations\citep{Chen2018}\citep{Kim2018}, the representations learned will not be perfect.
Motivated by this, we examine how the performance of HER (HAC with one level of hierarchy) and HAC (two levels of hierarchy) change as the representation of goals are systematically degraded from a ground truth baseline.
Specifically, we compare performance when the learned goal representation (mapping between observation and achieved goal) is rotated versions of the ground-truth, has additional factors, is noisy, or any combination of them. These are aberrations in goal spaces that are hand-designed or learned using $\beta$-variational auto-encoder~\citep{Higgins2017a} or its variant~\citep{Burgess2017}.
We show that performance in HAC is significantly affected by some of these modifications but not all.

\section{Related Work}
\label{sec:related_work}

Hindsight experience replay (HER)~\citep{Andrychowicz2017} has recently been shown to be effective in dealing with sparse rewards.
HER generalizes the value function to include a goal the agent must achieve.
It is a data augmentation technique where each transition is added to the replay memory multiple times, each with alternative goals in addition to the goal that was actually provided.
State transitions of the environment based on the actions taken are reformulated as goals that were successfully achieved.
As a result, successful and unsuccessful transitions occur at similar frequencies even if the real training data contains only unsuccessful or mostly unsuccessful transitions, thereby facilitating learning when external rewards are sparse.

Hierarchy~\citep{Dayan1993, Sutton1999} has been shown to be effective in dealing with long time scales, improving performance via temporal abstraction and learning skills transferable to new tasks ~\citep{Vezhnevets2017, Frans2018, florensa2017stochastic, nachum2018data, nachum2018near}.
Each level in the hierarchy operates on different time scales, allowing better reward propagation; thus the hierarchical agent is more effective at solving longer time scale problems.
Recently, Hierarchical Actor-Critic (HAC)~\citep{Levy2017} and HierQ~\citep{Levy2018} have examined combining HER and hierarchy.
The lowest level policy is trained with hindsight experience replay and acts in the environment.
Higher level policies act by choosing goals for the sub-policy directly below it.
The authors demonstrate the effectiveness of this architecture using agents with up to three levels.
Hierarchical Actor-Critic, which we explore in our work, is one among many previous works utilizing multiple policies operating at different time scales, the most well-known of which is Options~\citep{Sutton1999}.
~\citet{Bacon2016} introduced an end-to-end trained option critic algorithm.
Meta Learning Shared Hierarchies~\citep{Frans2018} introduced a master policy which selects from a discrete set of sub-policies every \(N\) time steps.
None of these however utilize advances made with hindsight experience replay or universal value functions~\cite{schaul2015universal}. 
Many others also explore hierarchical reinforcement learning agents~\citep{Schmidhuber91neuralsequence}~\citep{Parr1998}~\citep{Wiering1997} ~\citep{Vezhnevets1997}.
FeUdal networks~\citep{Vezhnevets2017} use two levels of hierarchy, one which select goals, and another which attempts to achieve those goals.
Goals set by the higher-level policy are interpreted by the lower-level policy as directions in a latent space.

Disentangled representations have been used to extract features from images using convolutional neural networks, for example~\citep{Jaderberg2017} and~\citep{Mnih2015_nature}.
A disentangled representation is one in which each latent variable ideally corresponds to a single ground truth factor of variation.
Recent work~\citep{Higgins2017b} has shown that agents using disentangled representations learned by a beta-variational auto-encoder encoding the observation can be used to learn policies with good transfer and have potential for improving sample efficiency and asymptotic performance.
To our knowledge, only~\citet{Higgins2017b} specifically examine the impact of using disentangled latent representations in reinforcement learning agents.
They study zero-shot transfer performance, but also show that disentangled representations provide better asymptotic performance than other representations on simple tasks.
A variational auto-encoder is used in addition to a recurrent network in~\cite{Ha2018} to create a world model on which a model-based controller is trained.
While this is outside of the scope of the hierarchical reinforcement learning algorithms we are concerned with, ~\cite{Ha2018} show the potential of using latent representations in reinforcement learning in general.
However, the advantages of disentangled versus entangled representations when using as a goal space is still unknown.

\section{Methods}
\label{sec:methods}
We experiment with an HER agent and a two level HAC agent as seen in Figure~\ref{fig:setup}A and provide goal spaces that mimic learned goal spaces in both these cases.
We use the Fetch robot arm platform reach and pick-and-place task in OpenAI gym~\citep{openai_gym} as seen in figure~\ref{fig:setup}B.
The reach task involves the robot arm moving its gripper to a random goal location, and the pick and place task involves picking an object from a random location and placing it at a random goal location.

The ground-truth goal space is defined in absolute x, y and z coordinates.
This was the goal space used in the original HAC and HER papers and is known to work well.
The ground-truth goal space is modified in the following controlled ways (definitions for $f_s$ as specified in Algorithm~\ref{algo:analysis_algo} in Appendix A): rotation of the ground truth factors (axis-alignment is typically expected of disentangled factors), adding noise to the factors, and, adding unused dimensions in the latent representation, represented by their mean of zero (motivated by variational auto-encoders trained when the number of latent factors are unknown). In addition, we also tried all combinations of these aforementioned modifications.
The space in which the master policy provides goals to the sub-policy, in other words, the action-space of the master policy (identified as $f_m$ in Algorithm~\ref{algo:analysis_algo} in Appendix A), was matched in its dimensionality to the modified goal-spaces of the sub-policy.
Performance with each of these modifications were analyzed with one-level and two-level hierarchies, denoted as HER and HAC respectively.

\section{Experimental Results}
\label{sec:result}

\begin{figure}[t]
\centering
\includegraphics[width=\textwidth]{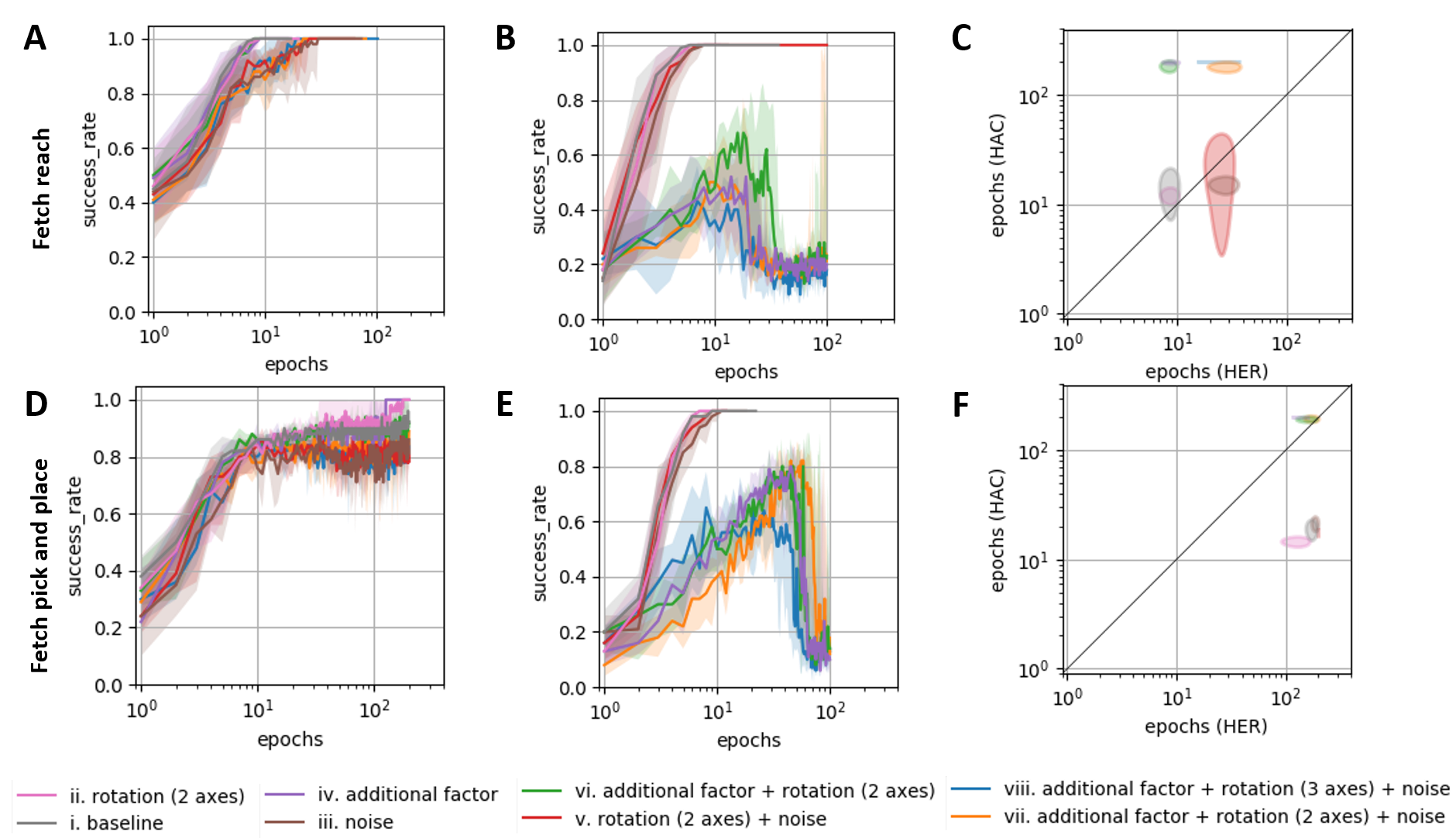}
\caption{(A) HER (1-level hierarchy) on Reach task (B) HAC (2-level hierarchy) on Reach task (C) HER vs HAC sample-efficiency comparison on Reach task (D) HER on Pick and Place task (E) HAC on Pick and Place task (F) HER vs HAC sample-efficiency comparison on Pick and Place task. Legend for all: (i) The baseline is the native goal space (perfect world coordinates). (ii) Rotation: two axes are rotated by pi/4. (iii) Noise: independent Gaussian noise is added to each factor. The signal to noise ratio is 18.75 dB. (iv) An additional, irrelevant factor is added to three ground truth factors. Its value is set to zero to simulate an unused dimension in a latent representation learned by a variational auto-encoder. (v) - (viii) Combinations of the individual perturbations. 100 trials were run for each goal space/algorithm pair in the case of fetch reach and 20 trials in the case of pick and place. The center line represents the median and the shaded areas represent the 25\% and 75\% quartiles.}
\label{fig:envs_perf_her_hac_compare}
\end{figure}

\subsection{Rotations}
\begin{figure}[t]
\centering
\includegraphics[width=1.\textwidth]{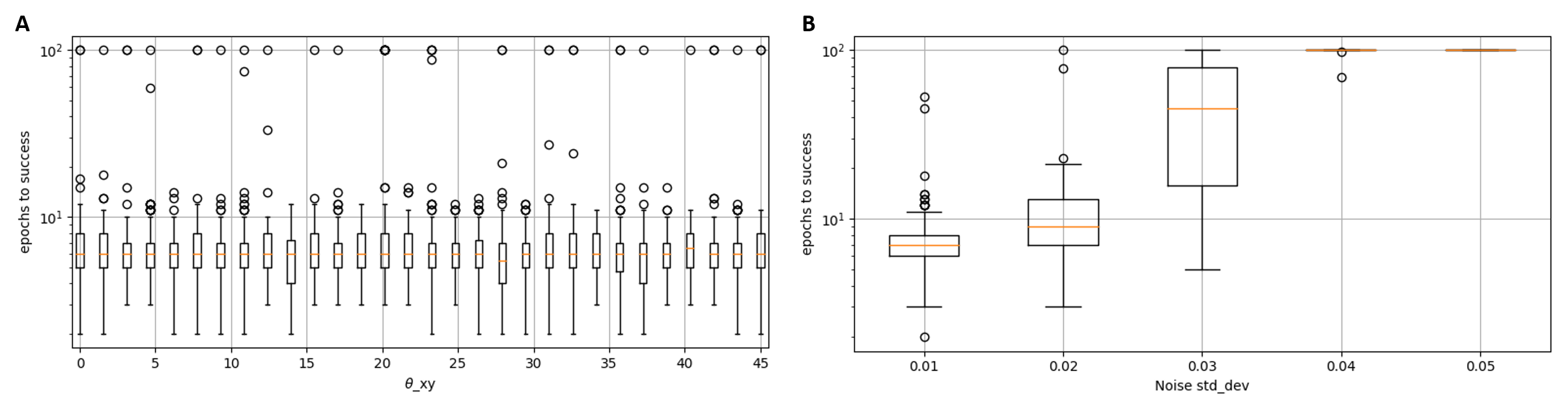}
\caption{Impact of entanglement and noise on performance of HAC in the reach task. [A]~Progressive rotations of the xy plane did not affect performance. Rotation by 45 degrees represents maximal entangling in a 2-D plane. Other pair-wise entanglements produce similar results, see Appendix B. [B]~As expected, increasing noise impairs HER learning and eventually makes the task unachievable. Gaussian noise is independently added to each factor, decreasing the SNR from 18.75dB to 6.71dB.}
\label{fig:rotate_scan_xy_noise}
\end{figure}

As seen in Figure~\ref{fig:envs_perf_her_hac_compare}, entangling the ground truth factors by way of rotations alone had no effect on sample efficiency in either HER or HAC for the reach and pick-and-place task. This is further supported by Figure~\ref{fig:rotate_scan_xy_noise}. While disentangled representations of factors have been postulated to be beneficial to downstream learning tasks in general~\citep{Higgins2017a, Chen2018, Kim2018}, and potentially beneficial in reinforcement learning in particular~\citep{Higgins2017a}, our results demonstrate that axis alignment is not critical in this case.

\subsection{Noise}
Adding a small amount of noise resulted in lower sample efficiency with HER but had negligible effect on HAC on both tasks (Figure~\ref{fig:envs_perf_her_hac_compare}). This was true for all combinations of goal space modifications that included noise. This is especially relevant when deploying on real hardware, where noise is likely to be present, both in manually designed and learned goal spaces. Gaussian noise is added with a mean of 0 and a standard deviation of 0.01. For reference, the master action space is constrained to +/- 0.1 in each factor, and the euclidean distance threshold used to determine if the current state achieves the desired goal is 0.1.

\subsection{Additional Factors}
Since latent representations are learned before the task is known, the representation may include features unnecessary to the task. In the case where multiple tasks must be achieved in the same episode, for example stacking two different blocks at a particular location, during the first half of the task when the first block must be moved to position, factors in the latent representation describing the second block can be seen as additional factors.

To study learning performance in the presence of additional factors we modified the goal space by adding a single constant 0.
HER performed no differently, but HAC took a dramatic hit.
HAC converged after 10 epochs without the addition, and did not converge after 100 with the additional factors.
Because the additional factor does not affect HER, we believe that the problem is not in the sub policy learning, but with the master policy learning that only a small subset of its possible action space is valid.
To test this hypothesis further, we removed the additional factor from the master policy goal space, leaving it only in the master policy action space and sub policy goal space.
HAC still did not converge after 100 episodes in any of the 10 trials we ran.
While initially counter intuitive, a similar issue was pointed out by the authors of HAC.
They found that the master policy would often have difficulty learning to select achievable sub goals for the sub policy.
By adding additional factors, learning to select achievable sub goals become significantly more difficult.
The authors introduced two modifications to their initial algorithm to address this.
First, a penalty when the master policy chooses goals which are not achievable,
and second, hindsight actions: the replay buffer of the master is populated with transitions in which the action taken by the master policy is replaced by the goals actually achieved by the sub-policy.
We implemented both unachievable sub-policy goal penalty and hindsight action memory, however they were not sufficient to overcome this particular modification of the goal space.

\section{Conclusion}
\label{sec:conclusion}
In this work, we highlight the question of what are desirable properties of goal spaces and how those properties may differ with a hierarchical architecture. Learned latent variable models will have imperfections, and our aim was to study these in a controlled manner in a relatively simple environment.
We experimented with two simulated robotics tasks in order to understand the effect various properties of the goal space on sample efficiency in the context of HER and a hierarchical version of HER, implemented as HAC. In these tasks, rotations of the goal space did not have an effect on the performance of either algorithm. However, an additional irrelevant factor dramatically affected the hierarchical agent while having no effect on single-level HER. This highlights the importance of the representation used for the used for the action space of the master policy. Finally, we see that noise affects HER more strongly than it affects a hierarchical model.


One approach to making HAC more robust is related to the master policy’s action space. In HAC, hindsight action memory is used to help the master policy more quickly learn to choose achievable actions. A penalty for selecting unachievable actions is then also added to ensure that the master policy does not get stuck repeatedly generating unachievable sub goals. ~\citet{Vezhnevets2017} discuss a similar issue, choosing to use directions in a learned state space as the goal space with the justification that goals are valid for a large part of the latent space. We find that by increasing the dimensionality of the goal space, HAC performs significantly worse, suggesting that there are still improvements to be had by more directly encouraging or enforcing the master policy to select achievable sub goals.
We are currently exploring different success functions that we believe would make HAC more robust to goal space aberrations.
Other future work involves extending these experiments to additional tasks in more complex environments. We envision building off these insights to design hierarchical agents which use learned, potentially disentangled, representations of goal spaces in complex visual environments.


\clearpage
\bibliography{paper}
\bibliographystyle{iclr2019_conference}

\section*{Appendix A}
\begin{algorithm}[H]
\DontPrintSemicolon
\SetAlgoLined
\KwResult{Learning curves for different modifications of the goal space}
Ground-truth goal space $G$, Modified goal spaces $F_m$ and $F_s$, Observation space $O$\;
Ground-truth factor function $g(O) \rightarrow G$\;
Desired goal provided by the environment $desired\_goal_{env} \in G$\;
Master-policy $\pi_m$, Sub-policy $\pi_s$\;
\For{$f_m(G) \rightarrow F_m$ and $f_s(G) \rightarrow F_s$ in Goal-space modification functions}{
\BlankLine
train $\pi_m$ and $\pi_s$ based on HAC algorithm with the following modifications:\newline
    $achieved\_goal_m = f_m(g(obs_{env}))$\newline
    $desired\_goal_{m} = f_m(desired\_goal_{env})$\newline
    $achieved\_goal_{s} = f_s(g(obs_{env}))$\;
}
\caption{Analysis approach based on simplification of architecture shown in figure~\ref{fig:setup}A}
\label{algo:analysis_algo}
\end{algorithm}

\section*{Appendix B}
\subsection*{Goal-space modification - Rotation}
\begin{figure}[H]
    \centering
    \includegraphics[width=1.\textwidth]{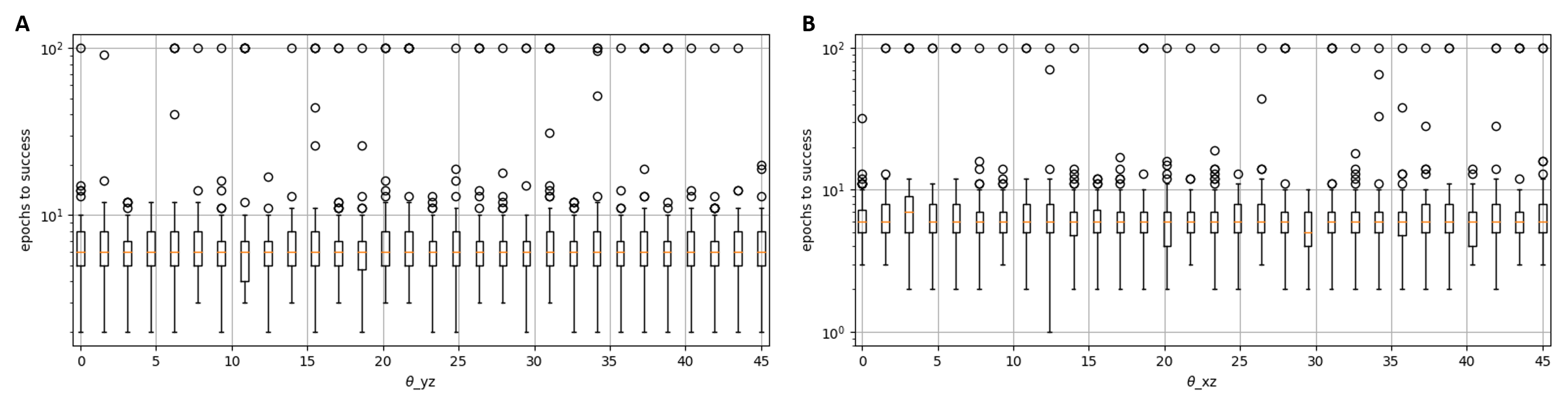}
    \caption{[A] and [B] show that performance as a consequence of rotating the 3-dimensional ground-truth goal space representation along the yz and xz plane respectively is indistinguishable while using HAC. Similar to rotations in the xy plane, 30 rotations in $[0, \pi4]$ were independently performed and 100 trials were executed for each angle.}
    \label{fig:sup_rotation}
\end{figure}

Since the target positions in reach tasks are in the xy co-ordinate space, the results section of the paper only showed rotations in that plane. Here in figure~\ref{fig:sup_rotation} we show that rotation in yz and xz planes show similar results, namely, learning in HAC is not compromised by goal spaces defined by rotation of ground truth goal space.

\end{document}